\documentclass[journal,10pt,onecolumn,draftclsnofoot]{IEEEtran}

\usepackage{cite,graphicx,subfig,multirow,array}

\usepackage{amsfonts,amssymb}
\usepackage[cmex10]{amsmath}
\usepackage{algorithmic}
\usepackage[ruled]{algorithm2e}	
\usepackage{epstopdf}
\usepackage{hyperref}
\usepackage{verbatim}

\usepackage[usenames,dvipsnames]{color}
\usepackage{soul}

\usepackage{colortbl}
\definecolor{mygray}{gray}{.9}

\DeclareMathOperator*{\argmax}{mean}

\begin{document} 
	
\pagenumbering{gobble}

\title{Accurate Light Field Depth Estimation \\with Superpixel Regularization over\\ Partially Occluded Regions}

\author{\IEEEauthorblockN{Jie~Chen, Junhui~Hou, Yun~Ni, and Lap-Pui~Chau}
	
\thanks{J. Chen, Y. Ni, and L.-P. Chau are with the School of Electrical \& Electronic Engineering, Nanyang Technological University, Singapore (e-mail: \{Chen.Jie, E150190,  ELPChau\}@ntu.edu.sg), J. Hou is with the Department of Computer Science, City University of Hong Kong (e-mail: jh.hou@cityu.edu.hk).}
}

	
\maketitle

\begin{abstract}

Depth estimation is a fundamental problem for light field photography applications. Numerous methods have been proposed in recent years, which either focus on crafting cost terms for more robust matching, or on analyzing the geometry of scene structures embedded in the epipolar-plane images. Significant improvements have been made in terms of overall depth estimation error; however, current state-of-the-art methods still show limitations in handling intricate occluding structures and complex scenes with multiple occlusions. To address these challenging issues, we propose a very effective depth estimation framework which focuses on regularizing the initial label confidence map and edge strength weights. Specifically, we first detect partially occluded boundary regions (POBR) via superpixel based regularization. Series of shrinkage/reinforcement operations are then applied on the label confidence map and edge strength weights over the POBR. We show that after weight manipulations, even a low-complexity weighted least squares model can produce much better depth estimation than state-of-the-art methods in terms of average disparity error rate, occlusion boundary precision-recall rate, and the preservation of intricate visual features.
\end{abstract}
\begin{keywords}
Light field, superpixel, partially occluded border region, weight manipulation
\end{keywords}
\section{Introduction} \label{sec_intro}

With the commercialization of light field cameras such as Lytro \cite{Ng2005} and Raytrix \cite{perwass2012single}, light field imaging has become a popular topic and is attracting extensive research and industrial attentions. The light field (LF) is a vector function that describes the amount of light propagating in every direction through every point in space \cite{lippmann1908la}. Compared with conventional 2D cameras, LF cameras can capture extra directional information for each light ray, and such information enables exciting applications such as refocusing, 3D scene reconstruction \cite {Kim2013, perra2016analysis}, material recognition \cite{wang20164d}, reflection/specularity removal \cite{ni2017reflection, tao2016depth}, and virtual/augment reality display \cite{Huang2015LFStereo}, to just name a few. For most of the potential applications, depth estimation is one of the most fundamental problems \cite{Kim2013, perra2016analysis,chen2015light,jia2016scene,ni2017reflection}, and its quality directly determines the performance of various subsequent applications. 


The sub-aperture images (SAI) decoded from the LF data provide a densely sampled multi-angle perspectives of a targeted scene \cite{Adelson1992,dansereau2013decoding}. Being theoretically similar to the problem of depth inference from stereo, the extremely narrow baseline between SAIs \cite{yu2013line} hinders direct application of stereo matching algorithms to the highly sub-pixel scenario. Current state-of-the-art LF depth estimation methods mostly focus on exploration of epipolar-plane image (EPI) features. Geometrical features of the EPI such as edge line slopes, spatial and angular variances \cite{tao2013depth} have been used to obtain a robust initial estimate. Higher level local and global structural reasoning have also been applied to resolve confusing areas with occlusions \cite{Wanner2012,wang2016occlusion,zhang2017light}. Efforts have also been made to overcome challenges of noisy and distorted \cite{williem2016robust, jeon2015accurate} LF inputs.

\begin{figure}[t]
	\centering
	\includegraphics[width=3.5in]{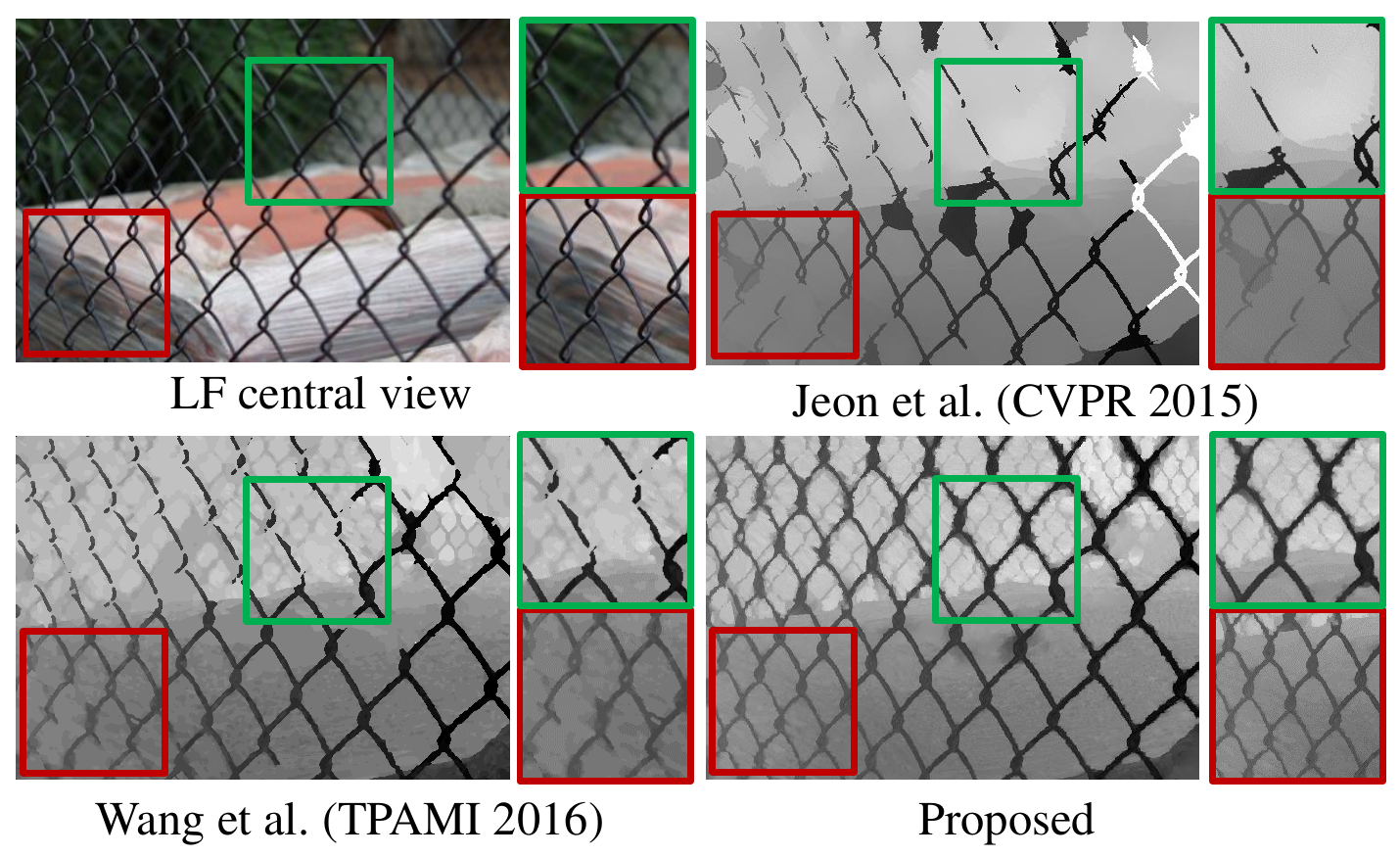}
	\caption{Comparison of estimated depth maps by different algorithms for a challenging scene with multiple occlusions from the Stanford Lytro LF Archive \cite{stanfordArchive}. Darker color indicates smaller distance to the camera. It can be seen that our method can capture fine details of occlusion boundaries, and works much better for intricate structures and overlaid occlusions (see the wired fence further behind). }
	\label{fig_dataset_teaser}
\end{figure}

\begin{figure*}[!t]
	\centerline{\subfloat{\includegraphics[width=6.4in]{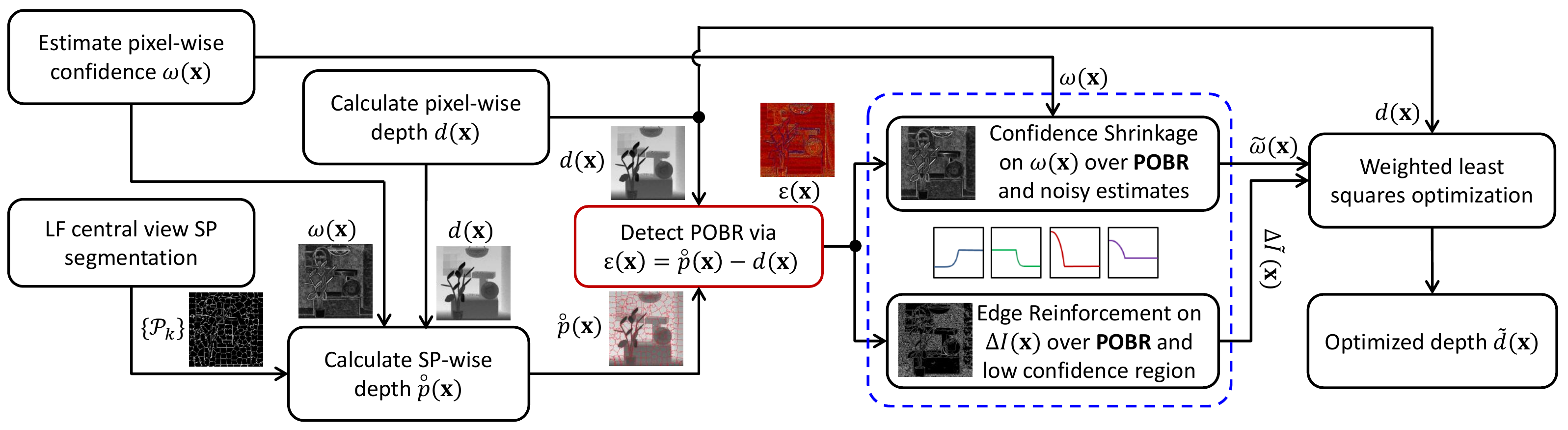}}}
	\caption{System flow chart for the proposed LF depth estimation algorithm.}
	\label{fig_systemFlow}
\end{figure*}

In spite of the significant improvements on the subject, occlusion remains one of the most difficult challenges especially when intricate occluding structures and complex scenes with multiple occlusions are involved. Color inconsistency over partially occluded regions causes existing methods to over-estimate their depth or even wipe out the structures. As shown in Fig. \ref{fig_dataset_teaser}, two state-of-the-art methods by Jeon \cite{jeon2015accurate} and Wang et al. \cite{wang2016occlusion} are respectively applied on a challenging LF scene from the Stanford Lytro LF Archive \cite{stanfordArchive}. Both methods obscure some occluding boundary details, such as the intricate wired fences near the camera, as well as the weak intensity ones at the back.

Unlike most current state-of-the-art methods, in which emphases are placed on crafting more robust cost terms for initial correspondence matching, we focus on regularizing the confidence map and the texture edge weight. The novelties and contributions our work are as follows:
\begin{enumerate}
\item we thoroughly analyze the geometrical causes for occlusion induced depth uncertainty in partially occluded border regions (POBR);
\item we propose to use superpixel based regularization to propagate local contextual information to the POBR to help resolve such uncertainty. We validated its efficiency in the detection of POBRs and the preserving of intricate occlusion boundaries through our experiments;
\item we propose a series of shrinkage and reinforcement operations over the initial depth label confidence and the texture edge weights, and integrate them into a final optimization framework. We show that after the weight/texture manipulation, even with a low-complexity weighted least squares model (as compared with the commonly used highly complex graph cut model), we can achieve better depth map estimation in terms of average disparity error rate, occlusion boundary precision-recall rate, and the preservation of intricate visual features.
\end{enumerate}

As can be seen in Fig. \ref{fig_dataset_teaser}, our method captures the fine details of occlusion boundaries, and works much better with intricate structures and overlaid occlusions.

The rest of the paper is organized as follows: Sec.~\ref{sec_related} introduces recently published works on this subject. Sec.~\ref{sec_occErr} gives a detailed analysis on how the uncertainty occurs for the POBR, and why it is challenging for current methods. Sec. \ref{sec_algo} gives a detailed introduction of the proposed algorithm: Sec. \ref{sec_algo_initial} explains how an initial pixel-wise depth map and initial confidence are estimated. Sec. \ref{sec_spDepthReg} introduces the role of SP regularization in POBR detection. Sec. \ref{sec_algo_conf} and Sec. \ref{sec_algo_edge} introduce how the confidence map and edge weights are modified, and Sec. \ref{sec_algo_final} shows how to fit them into the final optimization model. Comprehensive quantitative and qualitative evaluations and comparisons are carried out in Sec. \ref{sec_exp}. Finally, Sec. \ref{sec_conclusion} concludes the paper.

\section{Related Work} \label{sec_related}

The challenge to infer depth from LF images is similar to that of traditional stereo vision. However, instead of a pair of input images, the LF extends the disparity space to a continuous or multiple discrete ones. Bishop et al. \cite{Bishop2012} estimated the scene depth by iterative searching and filtering among multiple aliased views for the best correspondence match. Wanner et al. \cite{Wanner2012} used a structure tensor based on local gradients to estimate the direction of lines on EPI. Yu et al. \cite{yu2013line} studied geometric structures of 3D lines in ray space and encoded the line constraints to improve the reconstruction quality. Kim et al. \cite{Kim2013} proposed a scoring mechanism for all the hypothetical disparities for each scene point using dense light fields.

Extensive efforts have been made to craft an efficient cost term for initial depth estimation. Tao et al. \cite{tao2013depth} proposed to combine correspondence and defocus cost terms to calculate the depth, where they showed correspondence costs are more robust to regions of occlusions, while defocus costs are more robust to noisy regions with repeated texture. Jeon et al. \cite{jeon2015accurate} designed a cost-volume to increase sub-pixel accuracy of the depth estimation. 
Li et al. \cite{li2015continuous} used depth assisted segmentation to help solving a sparse linear system with two different affinity matrices. 
Williem et al. \cite{williem2016robust} proposed to improve robustness of data cost term in the presence of noise by improving the correspondence clue with a novel angular entropy metrics and an adaptive defocus response term. 
Navarro et al. \cite{Navarro2017robust} proposed a fusion framework among depth estimations from several pairs of two-view stereos for a unique and robust final estimation.
The geometrical relationships between SAIs have been explored by Heber et al. \cite{heber2014shape} by operating series of warpings over the SAIs into a low-rank matrix. Sparse coding \cite{johannsen2016sparse} and convolutional neural network \cite{heber2016convolutional} have also been used to improve robustness of the initial estimation.

Higher level structural reasoning has also been used to resolve the estimation uncertainties. Chen et al. \cite{chen2014light} proposed a bilateral metric to measure angular pixel patches' probability of occlusion by their similarity to the central view pixel. Wang et al. \cite{wang2016occlusion} proposed to model the boundary occlusion relationships by analyzing separately the two parts of the EPI divided by the central view. Zhu et al. \cite{zhu2017occlusion} proposed a complete occlusion model to guide the depth estimation for scenarios of multi-occlusions. All these methods can efficiently improve the estimation results over the POBR as compared with those without explicit occlusion analysis. However, their performances are limited either with added noise, or when the occluded regions are textureless. In contrast, our method combines the information of all pixels in a locally homogeneous superpixel region, and propagates through multiple ones. Therefore our method is more robust to noise and different texture patterns.

\begin{figure}[t]
	\centering
	\includegraphics[width=3.4in]{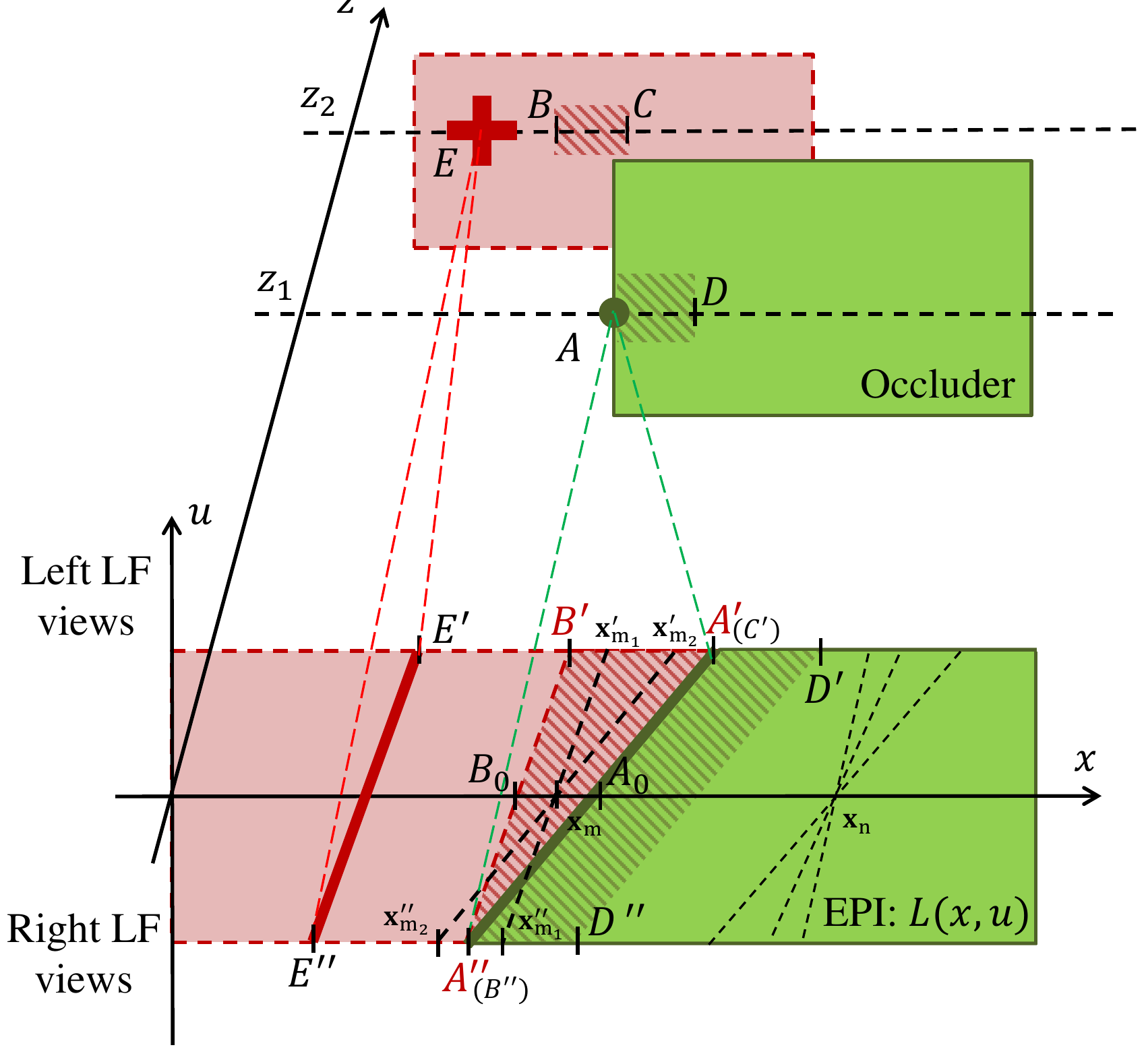}
	\caption{Illustration of the geometry of scene occlusion relationship and its pattern on the EPI. The triangular area $A'A''B'$ shaded in red is the POBR.}
	\label{fig_occEPI}
\end{figure}

\section{Occlusion Induced Uncertainty}\label{sec_occErr}

First of all, we re-visit the basic geometry of LF EPI, and establish some important consensus. In Fig. \ref{fig_occEPI}, suppose there are two objects placed at different distances from the camera. The nearer object in green whose distance is denoted as $z_1$ partially occludes the further object in red with the distance denoted as $z_2$. The EPI for such a scene configuration is shown below. $x$ and $u$ are the horizontal spatial and angular dimensions, respectively. 
The positive direction of $u$ corresponds to the left LF view angles. 

The two dotted lines over $z_1$ and $z_2$ are the actual object sections that form the EPI below. To be specific, the line $A'A''$ on the EPI is projected from point $A$ of the occluder $z_1$. Point $A'$ corresponds to $A$'s image location on the leftmost LF view, and $A''$ corresponds to its location on the rightmost view. The red shaded region $BC$ on $z_2$ is partially occluded by $z_1$: with zero occlusion for the leftmost LF view ($B$ projected to EPI as $B'$, $C$ projected as $C'$ overlapped with $A'$), and complete occlusion for the rightmost LF view ($C$ occluded, $B$ projected to EPI as $B''$ overlapped with $A''$). We define the triangular region $A'A''B'$ on the EPI as \textbf{partially occluded border region (POBR)}. The concept of POBR is very important for this work, and will be used throughout the paper. Note that region $A'A''D''D'$ is not considered as POBR, since it belongs to the occluding object $z_1$. 

\subsubsection{All lines on the EPI belong to the occluder} in Fig. \ref{fig_occEPI}, the line $A'A''$ is projected from point $A$ on the occluder, and its slope reflects the depth of $z_1$. The depth of $z_2$ is actually indicated by the slope of line $B'A''$ (in fact $B'B''$), but it is only a virtual line. A textured point $E$ on $z_2$ also forms a line denoted as $E'E''$ on the EPI, which is parallel to the line $B'A''$. However, this does not conflict with our observation, since $E$ is not occluded by any other objects. $z_1$ is an occluder at point $E$.

Depth can be calculated from both correspondence and defocus cues \cite{tao2013depth}. Although defocus cue works better over regions with noisy and/or repeated patterns and proves to be more robust over textureless regions, this advantage becomes negligible when the camera resolution is reasonably high, and when better imaging quality is available. In this work, we put our emphasis on the properties of the correspondence cue under occlusion influences.

The process of depth correspondence matching can be considered as comparing pixel intensity variances along different lines on the EPI \cite{Kim2013}. Consider a point  $\mathbf{x}_n$ of the central view ($u=0$) on the EPI in Fig. \ref{fig_occEPI}, multiple angular variances can be calculated along possible candidate lines. The one that produces the least variance is considered the best response, and its slope directly indicates the scene depth. Specifically, for a given slope for the depth $d'$ \cite{Ng2005}, angular variance is calculated as:
\begin{equation}\label{eq_angVar}
\sigma_{d'}(\mathbf{x}_n)^2= \frac{1}{N_u-1}\sum_{u'}[L(\mathbf{x}_n+u'(1-\frac{f_0}{d'}),u')-\bar{L}_{d'}(\mathbf{x}_n)]^2.
\end{equation}
Here $f_0$ is the currently focused depth; $N_u$ is the number of angular views along $u$; and $\bar{L}_{d'}(\mathbf{x}_n)$ is the angular mean along the slope line for $d'$:
\begin{equation}
\bar{L}_{d'}(\mathbf{x}_n)= \frac{1}{N_u}\sum_{u'}[L(\mathbf{x}_n+u'(1-\frac{f_0}{d'}),u')].
\end{equation}
Therefore, the depth estimated from correspondence cue is:
\begin{equation}
d(\mathbf{x}_n)= arg\min_{d'} \sigma_{d'}(\mathbf{x}_n).
\end{equation}

Correspondence matching proves to be efficient for most regions when there is no occlusion ambiguity. Occlusion edges (e.g., $A'A''$) and texture patterns (e.g., $B'B''$) give positive guide for correct matching. However, for the POBR, the ambiguity along occlusion boundaries could lead to a series of uncertainties and errors. 

\subsubsection{Most correspondence matching errors are \textbf{underestimation} for the POBR on the occluded object}
consider a point $\mathbf{x}_m$ of the central view located in the POBR $A'A''B'$ in Fig. \ref{fig_occEPI}. Since the point belongs to the occluded object $z_2$, the correct depth slope for $\mathbf{x}_m$ should be $\mathbf{x}'_{m_1}\mathbf{x}''_{m_1}$, which is parallel to the line $B'B''$. However, due to the interference from the occluding edge $A'A''$, the angular variance along $\mathbf{x}'_{m_1}\mathbf{x}''_{m_1}$ is most likely larger than that along $\mathbf{x}'_{m_2}\mathbf{x}''_{m_2}$, which goes parallel without intersection with $A'A''$. Consequently, the depth for the POBR region $z_2$ will be underestimated as that of its occluder $z_1$. Such an underestimation is universal for all POBRs.

To alleviate occluding edges' interference with the depth estimations of the POBR, Wang et al. \cite{wang2016occlusion} proposed to consider the upper and lower parts of the EPI separately. 
Between $A'A_0B_0B'$ and $A_0A''B_0$, only the one with smaller variance will be used for depth estimation. For the case of point $\mathbf{x}_m$, the intensity variance will be calculated along $\mathbf{x}'_{m_1}\mathbf{x}_m$ instead of the whole slope line $\mathbf{x}'_{m_1}\mathbf{x}''_{m_1}$. This method greatly reduces errors caused by the occluding edges. However, there is still no guarantee that the variance of $\mathbf{x}'_{m_1}\mathbf{x}_m$ is smaller than that of $\mathbf{x}'_{m_2}\mathbf{x}_m$, or any other half slop lines that do not intersect with $A'A_0$. This ambiguity is especially serious when the POBR region is textureless or when multiple occlusions exist.

To this end, we propose to solve such ambiguities for the POBR using superpixel (SP) based regularization. The SP can combine and propagate the local information from the occluded object to the POBR, and detect occlusion boundaries precisely. Based on the obtained depth estimate, a series of shrinkage and reinforcement operations will then be applied to the confidence map and edge weights. The final output can better resolve the correspondence uncertainties over the POBR.

\section{Proposed Algorithm}\label{sec_algo}

The system flow chart for our proposed algorithm is shown in Fig. \ref{fig_systemFlow}. With an initial pixel-wise depth estimation and its confidence map, the major procedures of the proposed algorithm include SP-wise depth estimation, POBR detection, and shrinkage/reinforcement operations over the confidence map and edge weights. Finally these components are combined into an unified optimization framework. Detailed descriptions for each component will be given in this section.

\subsection{Initial Pixel-wise Depth Estimation and Confidence}\label{sec_algo_initial}

To increase robustness, we apply a bilateral filter with local window diameter $W_{\sigma}$ on the angular variance $\sigma_{d'}$ as defined in Eq. (\ref{eq_angVar}), and the initial depth value will be obtained by solving:
\begin{equation}
d(\mathbf{x})= arg\min_{d'} C_{d'}(\mathbf{x}),
\end{equation}
\begin{equation}
C_{d}(\mathbf{x})= \sum_{x'\in W_{\sigma}}e^{\frac{-[L(\mathbf{x}',0)-L(\mathbf{x},0)]^2}{2\gamma^2}}\sigma_d(\mathbf{x}),
\end{equation}
where $\gamma$ is the bilateral filter parameter that controls the contribution of neighboring pixels with respect to their intensity similarity to $\mathbf{x}$. 

\begin{figure}[t]
	\centering
	\includegraphics[width=3.4in]{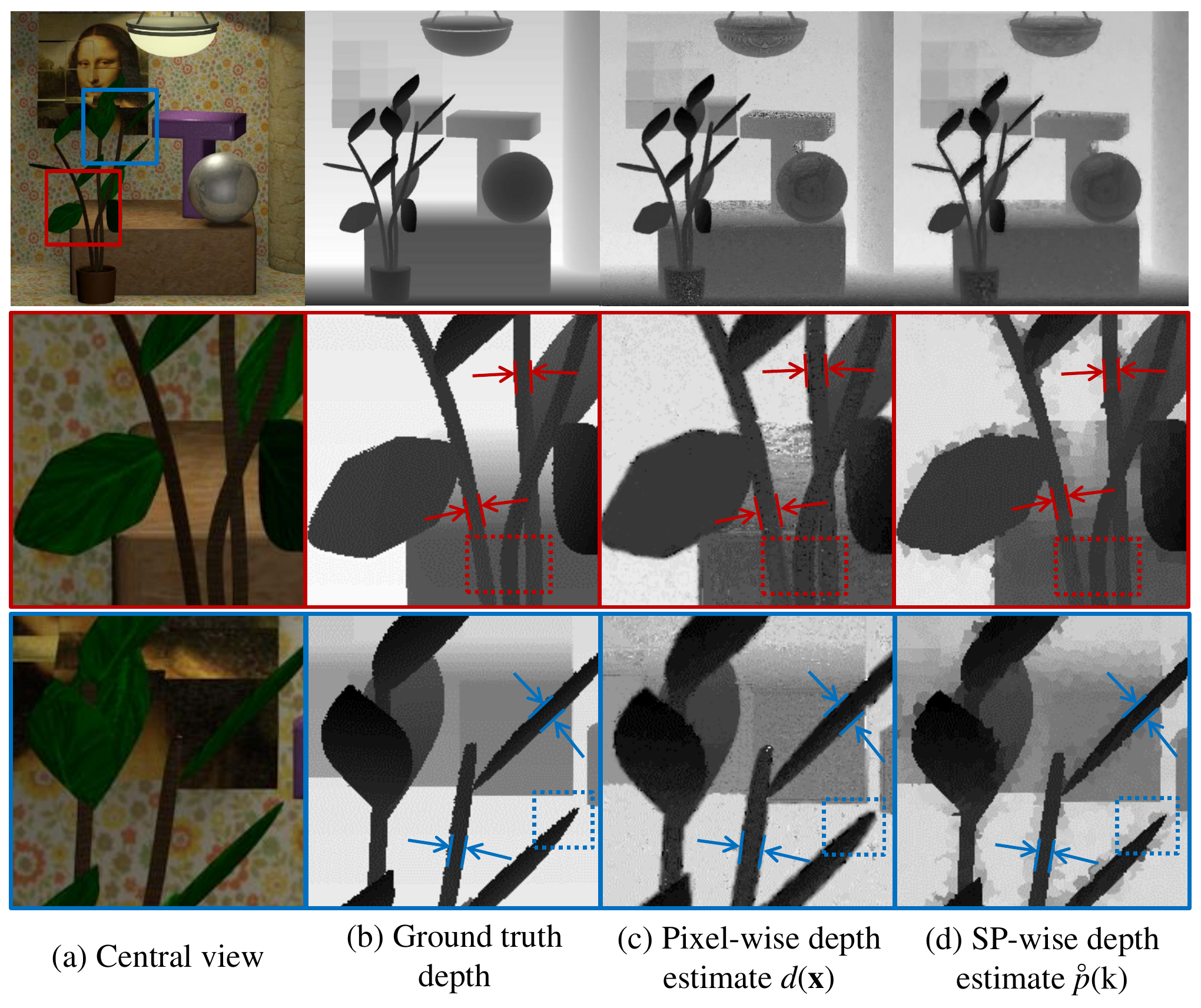}
	\caption{Comparison of ground truth depth for the LF \textit{Mona} in (b), pixel-wise depth estimate $d(\mathbf{x})$ in (c) and SP-wise depth estimate $\mathring{p}(\mathbf{x})$ in (d). 
		Notice the difference of stem width and leaf shapes among (b), (c) and (d).}
	\label{fig_depthConf_mona}
\end{figure}

We show the initial estimated depth map for the LF data \textit{Mona} from the HCI dataset \cite{Wanner2013} in Fig. \ref{fig_depthConf_mona}(c). As can be seen in the zoomed-in details in the second and third rows, the depth estimation for the occluding objects tend to be inaccurate along the edges. The bleeding effect is obvious: the bloated stem width, and the expanded leaf size. The depth for the POBR is underestimated.

We assign a depth estimation confidence to each $d(\mathbf{x})$ according to:
\begin{equation}
\omega(\mathbf{x})=  \mathcal{N}\{\dfrac{\argmax_{d'}C_{d'}(x)}{\min_{d''}C_{d''}(x)}\},
\end{equation}
which is the ratio between the mean and minimum variance among all possible depth candidates. $\mathcal{N}\{\cdot\}$ is a normalization operator that maps the confidence value to the range of [0,1].

\subsection{SP-wise Depth Estimation and POBR Detection}\label{sec_spDepthReg}

In order to propagate the local information from the occluded objects to the POBR, and thus reveal precise occlusion boundaries, we apply SP segmentation to the LF central view. The concept of SP has been widely used in various computer vision applications such as image segmentation \cite{ren2003learning} and object tracking \cite{wang2011superpixel}. It groups pixels into perceptually meaningful atomic regions. We employ the SLIC SP segmentation algorithm \cite{achanta2012slic} in our work, which is an iterative regional pixel clustering algorithm that clusters each pixel to an initiated center grid according to their respective normalized spatial and color distance. 

\begin{figure*}[!t]
	\centerline{\subfloat{\includegraphics[width=6.4in]{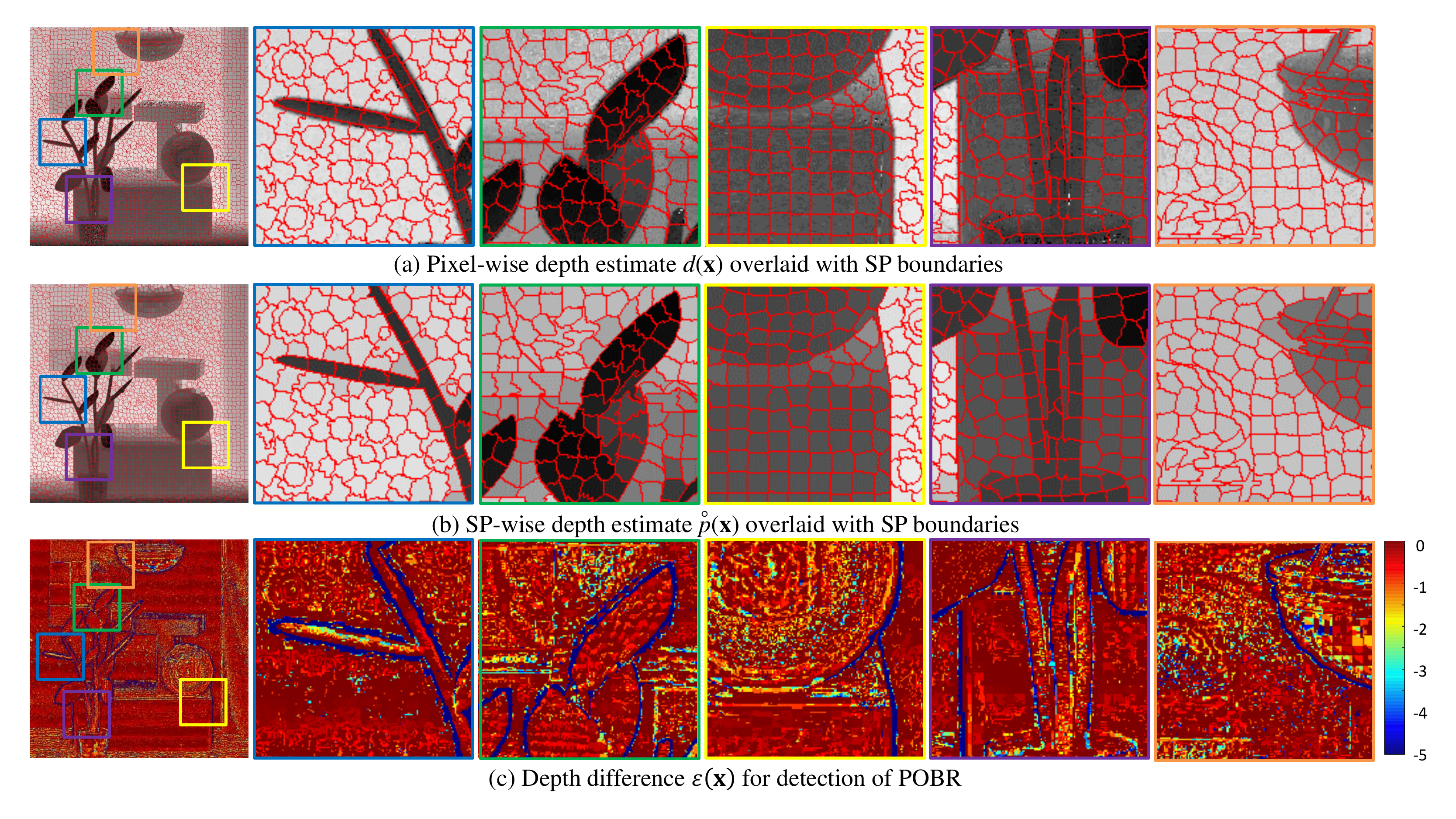}}}
	\caption{Depth estimates $d(\mathbf{x})$, $\mathring{p}(\mathbf{x})$, and $\varepsilon(\mathbf{x})$ map for the LF data \textit{Mona}. SP boundaries are overlaid in red. Regions in blue, green, yellow, and purple colored rectangles are zoomed in for comparison.}
	\label{fig_spOverlay}
\end{figure*}

Suppose we have a SP segmentation of the LF central view: $\{\mathcal{P}_k|k=1,2,...,n\}$, where $\mathcal{P}_k$ is the set of pixels that belongs to the superpixel $k$, and $n$ is the total number of SPs. Fig. \ref{fig_spOverlay}(a) shows the the SP segmentation output overlaid with the initial pixel-wise depth $d(\mathbf{x})$. Since SP segmentation is performed on the center view, the SP boundaries should adhere well to the object boundaries without interference from occlusion. As can be seen in Fig. \ref{fig_spOverlay}(a), the overlaid SP boundaries make the underestimated depth for the POBRs more obvious, i.e., the underestimated depth bleeds out of the object boundaries.

Based on the initial pixel-wise depth estimation $d(\mathbf{x})$, we propose to calculate a SP-wise depth map $p(k),k=1,2,...,n$, to make the estimations for each pixel more locally coherent. By enforcing all pixels within each SP to have the same depth value, we expect to propagate local contextual depth information to the POBR and suppress initial estimation noise.

\begin{figure}[t]
	\centering
	\includegraphics[width=3in]{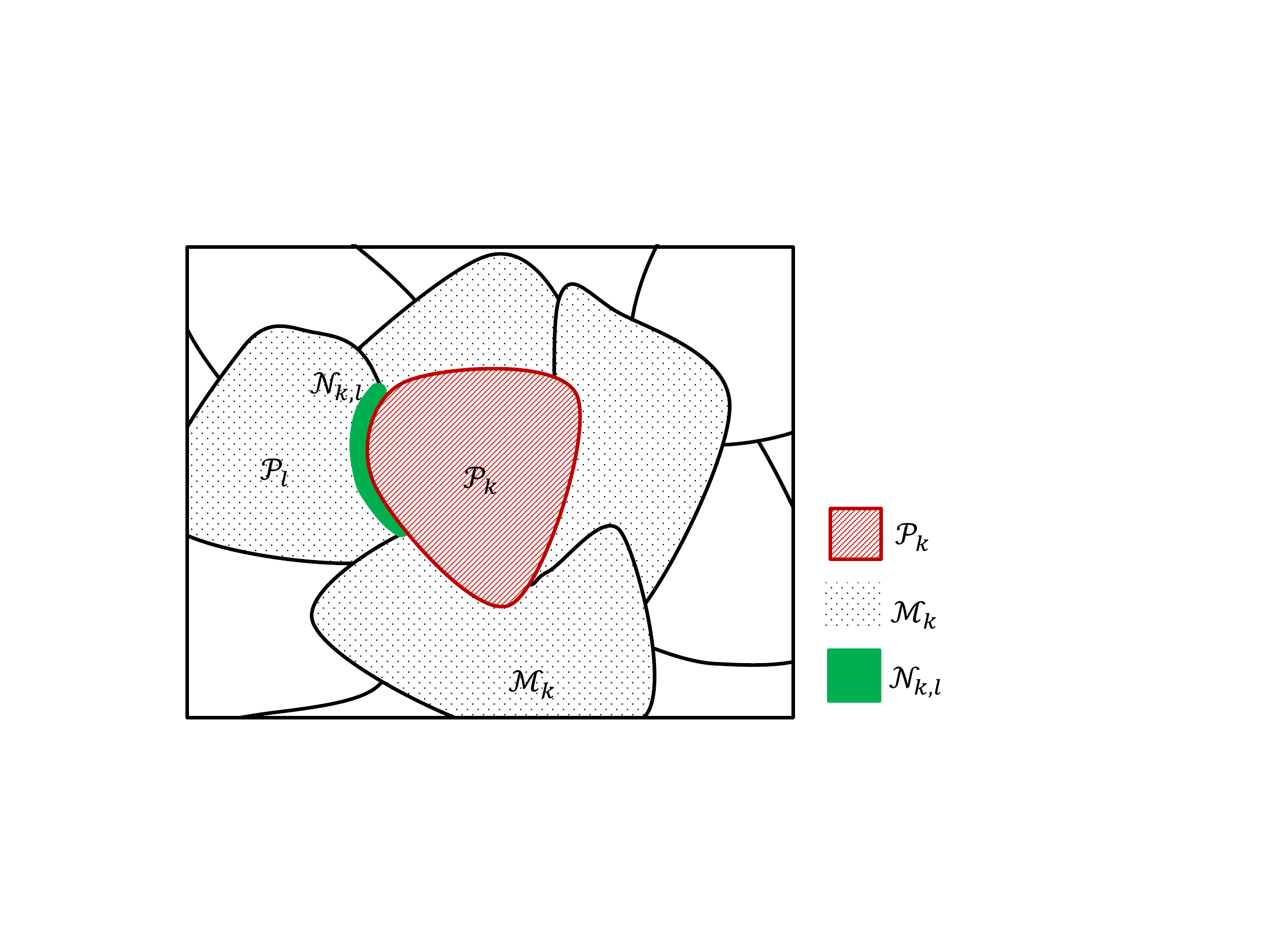}
	\caption{Illustration of SP neighborhood $\mathcal{M}_k$, and border pixels $\mathcal{N}_{k,l}$ defined for SP depth global regularization.}
	\label{fig_spnb}
\end{figure}

Let $\mathcal{M}_{k}$ denote the group of SPs that border the SP  $\mathcal{P}_{k}$, and $\mathcal{N}_{k,l}$ denote the set of boundary pixels in $\mathcal{P}_{l}$ ($\mathcal{P}_{l}\in\mathcal{M}_{k}$) which directly borders $\mathcal{P}_{k}$. Fig. \ref{fig_spnb} gives a visual illustration of $\mathcal{M}_{k}$ and $\mathcal{N}_{k,l}$ with respect to the location of $\mathcal{P}_{k}$. To obtain the depth value $p(k)$ for each superpixel, we minimize the following energy function:
\begin{equation}\label{eq_spReg}
\sum_{\mathbf{x}\in\mathcal{P}_{k}} \omega(\mathbf{x})|| p(k)- d(\mathbf{x})||_2+ \lambda\sum_{l\in\mathcal{M}_{k}}\sum_{\mathbf{y}\in{\mathcal{N}_{k,l}}}\frac{||p(k)-p(l)||_2}{||\triangledown I(\mathbf{y})||_1}.
\end{equation}

The first term in Eq. (\ref{eq_spReg}) forces $p(k)$ to be close to the initial pixel-wise depth estimation $d(\mathbf{x})$ for each pixel in the SP ( $\mathbf{x}\in\mathcal{P}_k$), up to a strength modulated by the initial confidence $\omega(\mathbf{x})$. For the second term, the numerator forces the SP-wise depth estimate $p(k)$ to be close to its neighbors $p(l)$ ($l\in\mathcal{M}_k$), and the denominator reduces the strength of the constraint when the sum of absolute gradients $||\triangledown I(\mathbf{y})||_1$ of the border pixels $\mathbf{y}\in\mathcal{N}_{k,l}$ is large.

Now we propagate the depth value of $p(k)$ to all pixels $\mathbf{x}\in \mathcal{P}_k$, and repeat for all $k=1,2,...,n$. The resulting SP-wise depth map $\mathring{p}(\mathbf{x})$ for the LF image \textit{Mona} is shown in Fig. \ref{fig_depthConf_mona}(d), where it can be seen that the width of the plant stem and the leaf size in $\mathring{p}(\mathbf{x})$ are almost close to the ground truth. Fig. \ref{fig_spOverlay}(b) shows $\mathring{p}(\mathbf{x})$ overlaid with SP boundaries. Compared with the initial pixel-wise depth $d(\mathbf{x})$ in Fig. \ref{fig_spOverlay}(a), the SP-wise depth is much less noisy and more regularized in POBRs. Besides, the SP-wise depth boundaries adhere to the true object boundaries well. 

\begin{figure}[t]
	\centering
	\includegraphics[width=3.4in]{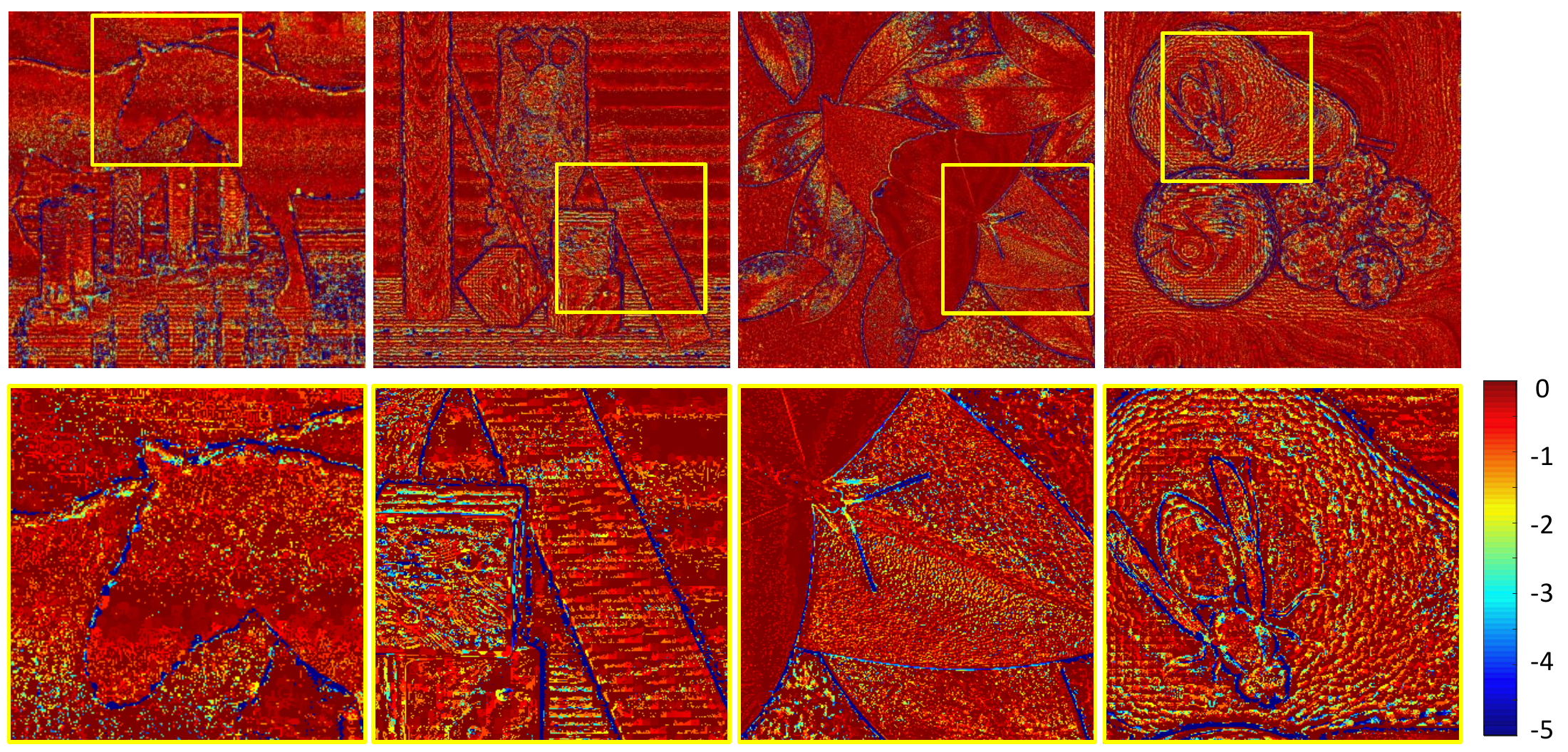}
	\caption{Calculated $\varepsilon(\mathbf{x})$ maps for the LF data \textit{horse}, \textit{Buddha2}, \textit{Papillon}, and \textit{stillLife}. Second row are zoomed-in segments indicated by yellow boxes on the first row. The POBRs are correctly registered as dark blue (small negative values). The occlusion boundary details are well captured.}
	\label{fig_epsilonMap}
\end{figure}

As analyzed in Sec. \ref{sec_occErr}, underestimation is universal for the pixel-wise depth map $d(\mathbf{x})$ in the POBRs. Since the SP-wise depth map $\mathring{p}(\mathbf{x})$ can propagate the correct depth from the occluded surface to the correct occlusion boundary, their subtraction
\begin{equation}
\varepsilon(\mathbf{x})= d(\mathbf{x})-\mathring{p}(\mathbf{x})
\end{equation}
can give valuable information on the location of POBR. In this paper, we propose to use the negativity of $\varepsilon(\mathbf{x})$ to indicate the probability of POBR. 

\begin{figure}[t]
	\centering
	\includegraphics[width=3.4in]{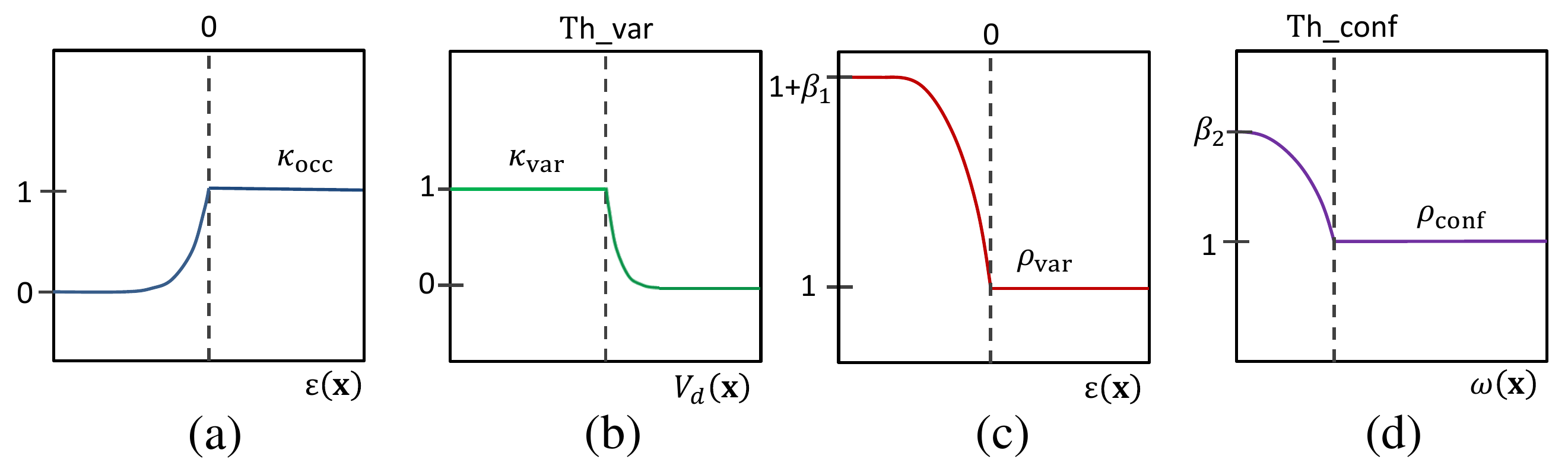}
	\caption{Curve plots for the weight manipulation functions $\kappa_\text{occ}$, $\kappa_\text{var}$, $\varepsilon_\text{occ}$, and $\varepsilon_\text{conf}$.}
	\label{fig_manipulateWeight}
\end{figure}

Fig. \ref{fig_spOverlay}(c) shows the $\varepsilon(\mathbf{x})$ map for the LF data \textit{Mona}, in which darker blue color indicates smaller negative values in $\varepsilon(\mathbf{x})$ (in units of depth labels). As can be seen, the POBRs have been correctly registered as dark blue color; and the well aligned occlusion boundaries from $\mathring{p}(\mathbf{x})$ also help to ensure precise alignment of occlusion boundaries for $\varepsilon(\mathbf{x})$. More $\varepsilon(\mathbf{x})$ maps are shown in Fig. \ref{fig_epsilonMap} for the LF data \textit{horse}, \textit{Buddha2}, \textit{Papillon}, and \textit{stillLife}, where we can see the POBRs are correctly detected and occlusion boundaries are well aligned for all cases.

Based on the estimated $\varepsilon(\mathbf{x})$ map, in the following two subsections, we introduce a series of shrinkage/reinforcement operations on the confidence map $\omega(\mathbf{x})$ and the weights of the texture edges over the POBR, which we then integrate into our final depth optimization framework.

\subsection{Label Confidence and Edge Strength Manipulation}\label{sec_algo_conf}

\subsubsection{Confidence shrinkage over POBRs}
we define the shrinkage function $\kappa_\text{occ}(\mathbf{x})$ to refine the initial confidence map $\omega(\mathbf{x})$ over POBRs:
\begin{equation}\label{equ_ThreshDiff}
\kappa_\text{occ}(\mathbf{x})=\begin{cases}
\dfrac{2}{1+e^{-\varepsilon(\mathbf{x})}}& \text{$\varepsilon(\mathbf{x})<0$}\\
1& \text{$\varepsilon(\mathbf{x})\geq 0$}.
\end{cases}\\
\end{equation}
The curve plot for $\kappa_\text{occ}(\mathbf{x})$ is shown in Fig. \ref{fig_manipulateWeight}(a). The confidence $\omega(\mathbf{x})$ is shrank for the POBR ($\varepsilon(\mathbf{x})<0$), since its depth estimation is highly unreliable. The confidence is shrank more for smaller negative values of $\varepsilon(\mathbf{x})$. 

\subsubsection{Confidence shrinkage over uncertain noisy regions}
we also shrink the initial confidence $\omega(\mathbf{x})$ over the noisy initial depth estimations according to:
\begin{equation}\label{equ_ThreshVar}
\kappa_\text{var}(\mathbf{x})=\begin{cases}
\dfrac{2}{1+e^{(V_d(\mathbf{x})-\varGamma_v})}& \text{$V_d(\mathbf{x})> \varGamma_v$}\\
1& \text{$V_d(\mathbf{x}) \leq \varGamma_v$},
\end{cases}\\
\end{equation}
where $V_d(\mathbf{x})$ is the spatial variance of $d(\mathbf{x})$ in a small local widow, and $\varGamma_v$ is the variance threshold. Noisy estimations are usually over textureless noisy surfaces or highly over saturated regions. The shrinkage of confidence for these regions helps to regulate noisy predictions. The curve plot for $\kappa_\text{var}(\mathbf{x})$ is shown in Fig. \ref{fig_manipulateWeight}(b).

Finally, a refined label confidence map denoted as $\tilde{\omega}(\mathbf{x})$ is obtained:
\begin{equation}
\tilde{\omega}(\mathbf{x})= \omega(\mathbf{x})\cdot \kappa_\text{occ}(\mathbf{x}) \cdot \kappa_\text{var}(\mathbf{x}).
\end{equation}

\begin{figure}[t]
	\centering
	\includegraphics[width=3.4in]{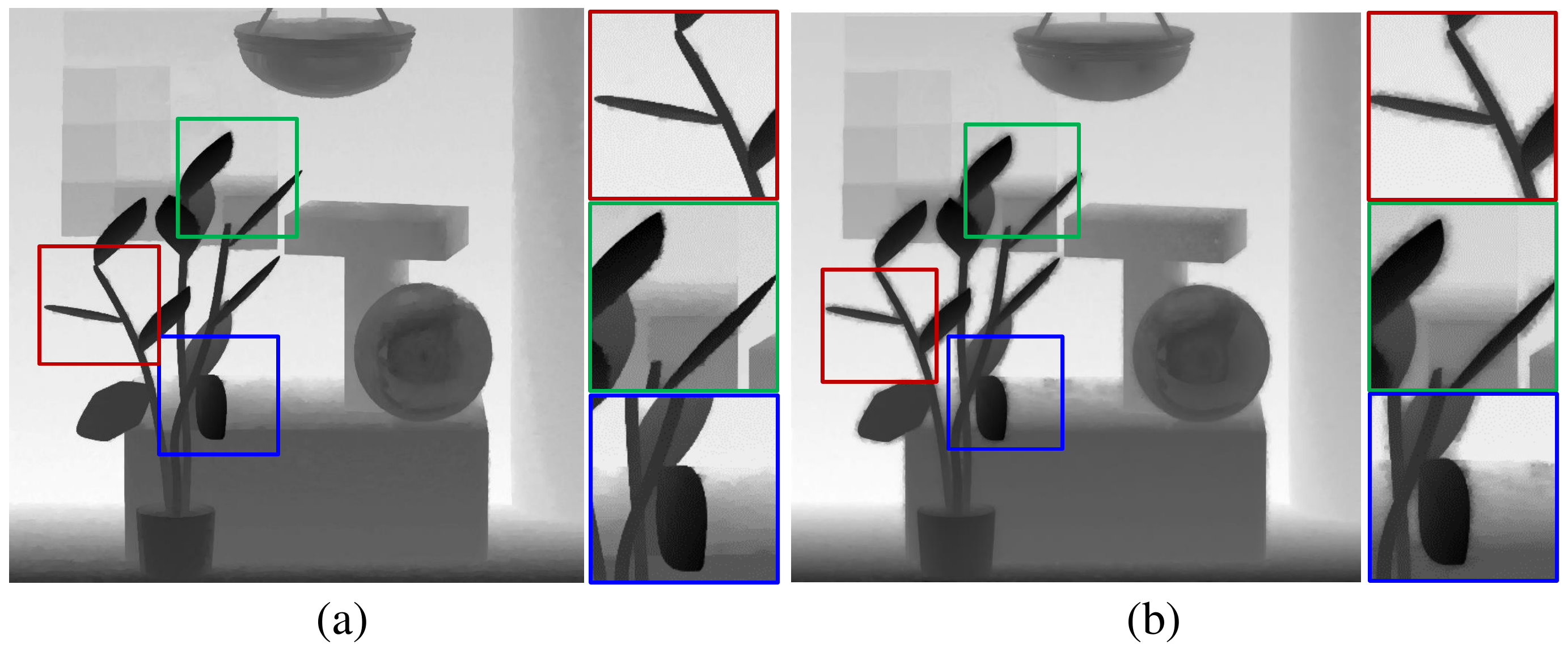}
	\caption{Demonstration of the effectiveness of weight manipulations. (a) is the outcome according to Eq. \ref{eq_finalDepth} with confidence map and edge strength manipulations as explained in Sec. \ref{sec_algo_conf} and \ref{sec_algo_edge}, while (b) is the output from the same model but with $\kappa_\text{occ}(\mathbf{x})$, $\kappa_\text{var}(\mathbf{x})$, $\rho_\text{occ}(\mathbf{x})$, and $\rho_\text{conf}(\mathbf{x})$ all set to 1.}
	\label{fig_compWeightManipulate}
\end{figure}

\subsection{Edge Strength Manipulation}\label{sec_algo_edge}
In our global regularization framework, depth discontinuities will be encouraged to occur over regions containing strong edges.
\subsubsection{Edge reinforcement over POBRs}
we increase the strength of edges in the POBR, such that depth discontinuities are encouraged to occur over these edges. Stronger edges also make the depth boundaries align more consistently with RGB edges. To this end, we introduce the following operation:
\begin{equation}\label{equ_edgeEnforce1}
\rho_\text{occ}(\mathbf{x})=\begin{cases}
1+ \beta_1\cos(\frac{\pi}{2}\kappa_\text{occ}(\mathbf{x}))& \text{$\varepsilon(\mathbf{x})<0$}\\
1& \text{$\varepsilon(\mathbf{x})\geq 0$}
\end{cases}\\
\end{equation}
The curve plot for $\rho_\text{occ}(\mathbf{x})$ is shown in Fig. \ref{fig_manipulateWeight}(c). This operation proves to be efficient especially for
occlusion boundaries with weak or blurred intensity gradients.

\subsubsection{Edge reinforcement over low confidence regions}
we also increase edge weights for regions with extremely low initial confidence so that depth discontinuities can be more flexible over these edges.
\begin{equation}\label{equ_edgeEnforce2}
\rho_\text{conf}(\mathbf{x})=\begin{cases}
1+ \beta_2\cos(\frac{\pi}{2}\omega(\mathbf{x}))& \text{$\omega(\mathbf{x})<\varGamma_c$}\\
1& \text{$\omega(\mathbf{x})\geq \varGamma_c$},
\end{cases}\\
\end{equation}
where $\varGamma_c$ is the threshold for the low confidence region. The curve plot for $\rho_\text{conf}(\mathbf{x})$ is shown in Fig. \ref{fig_manipulateWeight}(d).

\subsection{Final Depth Optimization} \label{sec_algo_final}

We integrate the refined label confidence map and edge strength weights into a global regularization framework:
\begin{align}\label{eq_finalDepth}
\min_{\hat{d}(\mathbf{x})} & \sum_{\mathbf{x}} \tilde{\omega}(\mathbf{x})|| \hat{d}(\mathbf{x})-d(\mathbf{x})||_2+\\\notag &\eta\sum_{\mathbf{x}}\sum_{\mathbf{y}\in{\Omega_{\mathbf{x}}}}\frac{||\hat{d}(\mathbf{x})-\hat{d}(\mathbf{y})||_2}{|| I(\mathbf{x})-I(\mathbf{y})||_1\cdot\rho_\text{occ}(\mathbf{x})\cdot\rho_\text{conf}(\mathbf{x})}.
\end{align}
The first term is the data fidelity term which is re-weighted with the refined confidence map $\tilde{\omega}(\mathbf{x})$. The second term is the smoothness term, with edge strength reinforced with $\rho_\text{occ}(\mathbf{x})\cdot\rho_\text{conf}(\mathbf{x})$. $\eta$ is the parameter that controls trade-off between the fidelity and the smoothness constraints. $\Omega_{\mathbf{x}}$ denotes the four nearest neighbors of $\mathbf{x}$ in the central view.

Eq. (\ref{eq_finalDepth}) is a weighted least squares problem, which can be solved with high efficiency as a large system of linear equations \cite{strutz2010data}. We use this model instead of the highly complex graph cut model, owing to the advantage brought about by the regularized confidence map $\tilde{\omega}(\mathbf{x})$ and edge strength $\rho_{\text{(occ)}(\mathbf{x})}\cdot\rho_{\text{(conf)}(\mathbf{x})}$. We show the effectiveness of such manipulations in Fig. \ref{fig_compWeightManipulate}, where Fig. \ref{fig_compWeightManipulate}(a) shows the estimated depth map via Eq. (\ref{eq_finalDepth}), and Fig. \ref{fig_compWeightManipulate}(b) shows the depth output from Eq. (\ref{eq_finalDepth}) with $\kappa_\text{occ}(\mathbf{x})$, $\kappa_\text{var}(\mathbf{x})$, $\rho_\text{occ}(\mathbf{x})$, and $\rho_\text{conf}(\mathbf{x})$ all set to 1. We can see that confidence and edge strength manipulations are important for the low complexity weighted least squares model to produce good results, especially for the POBRs.

\begin{table}[t]
	\begin{center}
		\caption{Percentage of disparity estimation error larger than 0.1 (pixel) for different methods on the HCI dataset (in \%).} 
		\label{tbl_quantEval}
		\begin{tabular}{|c|>{\centering\arraybackslash}m{1.35cm}|>{\centering\arraybackslash}m{1.2cm}|>{\centering\arraybackslash}m{1.2cm}|>{\centering\arraybackslash}m{1.2cm}|>{\centering\arraybackslash}m{1.55cm}|>{\centering\arraybackslash}m{1cm}|}
			\hline 
			&Wanner et al. \cite{Wanner2012} &Jeon et al. \cite{jeon2015accurate} &Wang et al. \cite{wang2016occlusion} &Proposed \\ 
			\hline \rowcolor{mygray}
			\textit{Buddha} &2.0 &6.4 &1.8 &2.0 \\ 
			\textit{Buddha2} &7.3 &4.8 &6.9 &4.5 \\ \rowcolor{mygray}
			\textit{Mona} &4.3 &6.0 &5.0 &2.6 \\
			\textit{Papillon} &13.7 &9.1 &6.7 &2.6 \\ \rowcolor{mygray}
			\textit{StillLife} &10.8 &13.6 &5.0 &5.6 \\
			\textit{Horses} &24.8 &10.6 &3.7 &4.3 \\ \rowcolor{mygray}
			\textit{Medieval} &8.7 &3.4 &3.6 &2.3 \\\hline
			\textbf{Average} &\textbf{10.2} &\textbf{7.70} &\textbf{4.67} &\textbf{3.55} \\\hline
		\end{tabular}
	\end{center}
\end{table}

\section{Experimental Results} \label{sec_exp}

We evaluate our algorithm and compare it with current state-of-the-art methods, i.e., the methods by Wanner et al. \cite{Wanner2012}, Jeon et al. \cite{jeon2015accurate}, and Wang et al. \cite{wang2016occlusion}. Parameters involved in these algorithms were set to different values for different datasets/scene according to suggestions from their respective authors. For our algorithm, we set $W_{\sigma}=5$, $\varGamma_v=0.3$, $\varGamma_c=0.1$, $\beta_1=5$, and $\beta_2=2$. The average number of pixels for each SP is set to be 50 for synthetic LF data, and 25 for LF data taken with Lytro Illum camera. $\eta$ is set to be within the range $[0.01,0.05]$ for different data.

We quantitatively evaluate the disparity estimation error and the occlusion boundary precision-recall rate for each algorithm on the HCI LF benchmark dataset \cite{Wanner2013}, which provides ground truth scene depth/disparity. Qualitative visual comparison of the estimated depth maps by different algorithms are carried out on the dataset provided by Wang et al. \cite{wang2016occlusion}, and some challenging data from the Stanford Lytro LF Archive \cite{stanfordArchive}. We also test several LF data captured by ourselves using a Lytro Illum camera.

\subsection{Comparison of LF Disparity Estimation Error} \label{sec_quantEval}

We quantitatively evaluate the disparity map estimation errors based on the HCI LF benchmark dataset \cite{Wanner2013}. With provided ground truth depth map and meta data for each synthetic LF scene, the ground truth disparity map can be directly calculated. The disparity estimations from each algorithm are compared against the ground truth, and we calculate the percentage of pixels with estimation error larger than 0.1 pixels. TABLE \ref{tbl_quantEval} lists corresponding results. As can be seen, our algorithm produces the best estimations with the smallest average disparity error rate of 3.55\%.

\begin{figure}[t]
	\centering
	\includegraphics[width=3.5in]{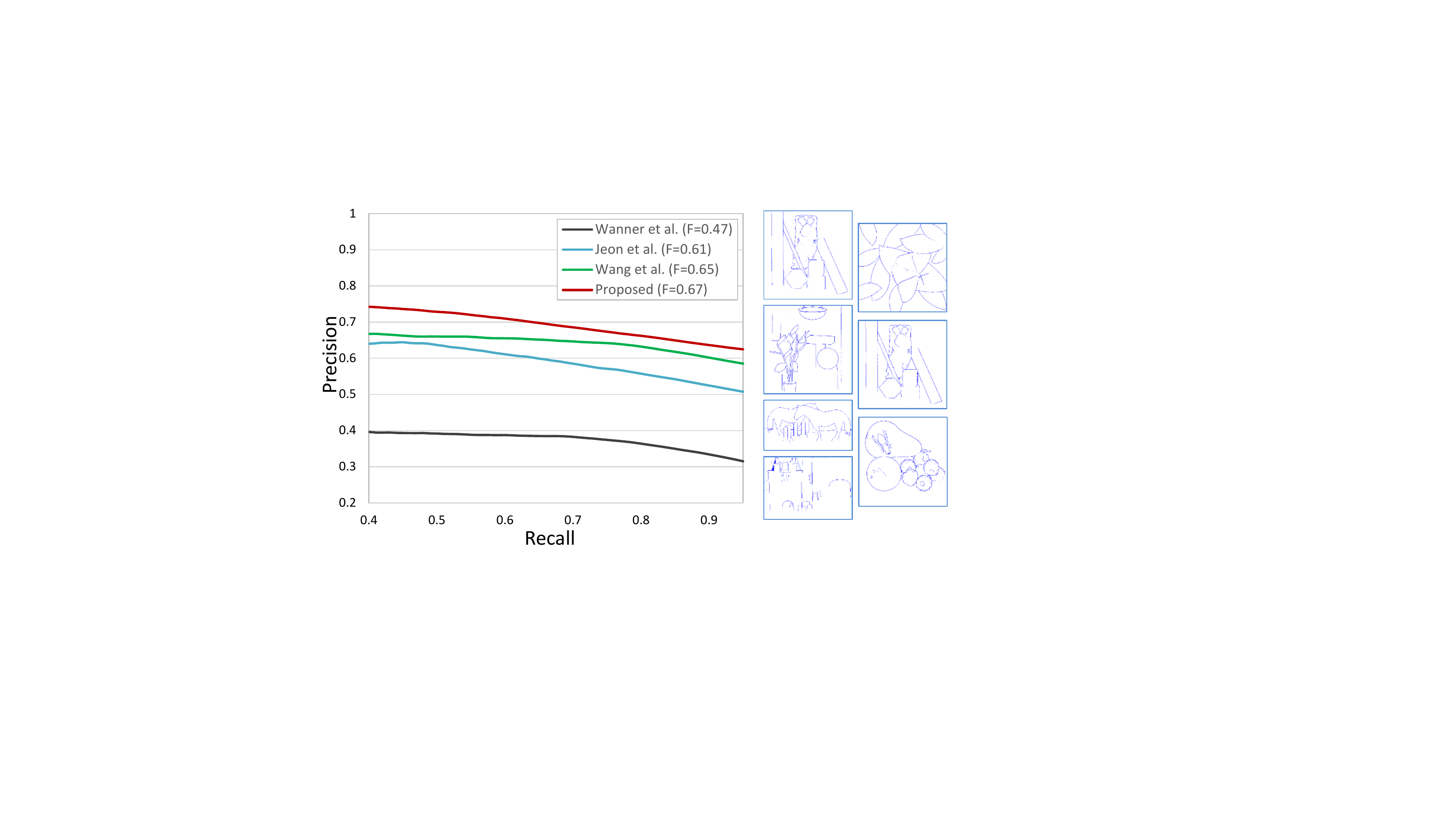}
	\caption{Comparison of the PR curves of occlusion boundaries by different algorithms on the HCI LF dataset. The occlusion boundary ground truth for each data are shown on the right.}
	\label{fig_curvePR}
\end{figure}

\begin{figure*}[!t]
	\centerline{\subfloat{\includegraphics[width=6.4in]{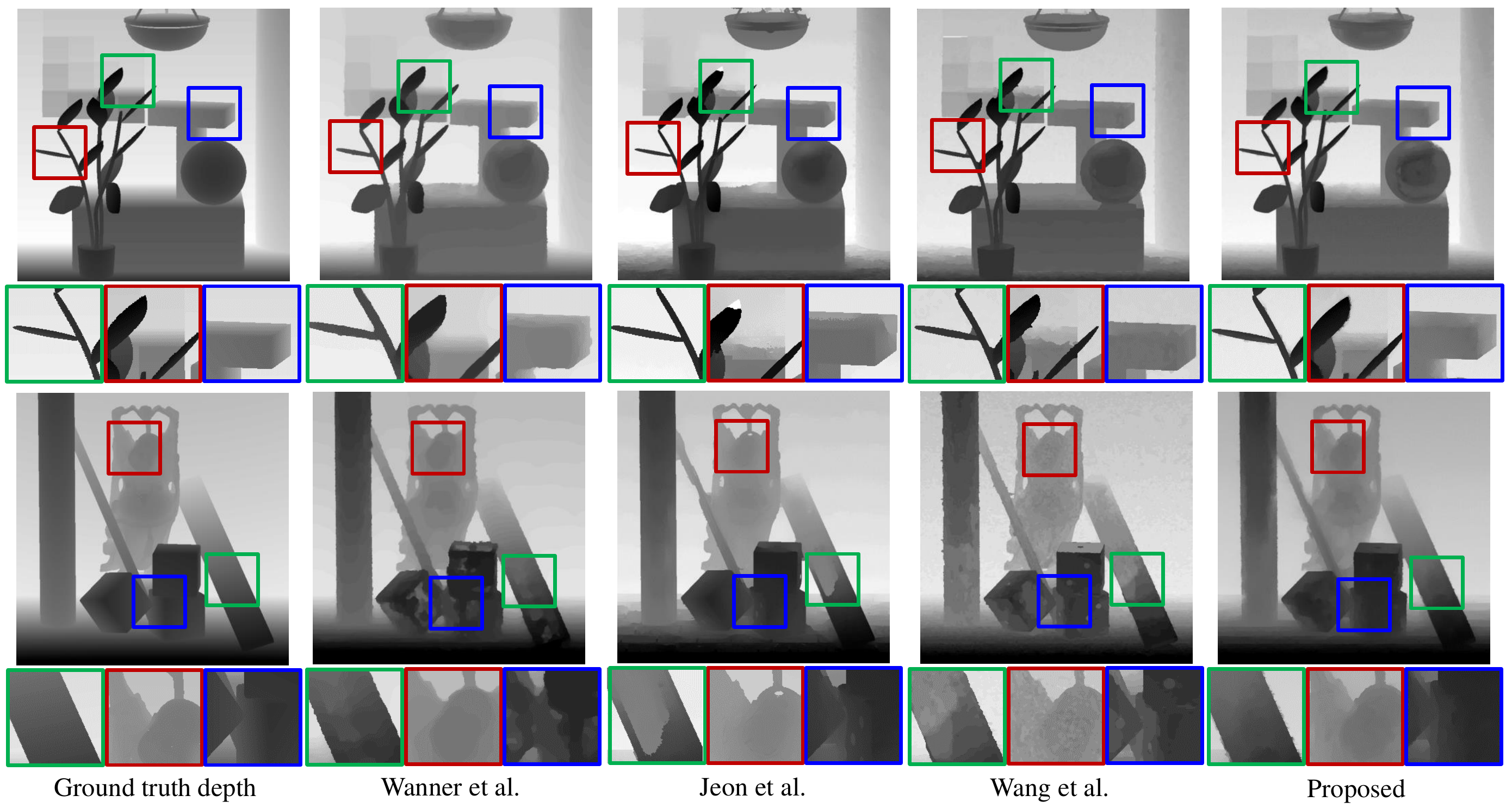}}}
	\caption{Visual comparison of estimated depth maps by Wanner et al. \cite{Wanner2012}, Jeon et al \cite{jeon2015accurate}, Wang et al \cite{wang2016occlusion} and our proposed method for the LF data \textit{Mona} and \textit{Buddha2} from the HCI dataset \cite{Wanner2013}.}
	\label{fig_dataHCI}
\end{figure*}

\subsection{Comparison of the precision-recall curve of occlusion boundaries} \label{sec_occlusionPR}

We compare the precision-recall (PR) curves of the occlusion boundaries between different algorithms. We calculated the gradients of the ground truth depth map for each synthetic LF data in the HCI dataset \cite{Wanner2013}. Then we manually found an appropriate threshold for each data to segment the depth gradients and got the ground truth occlusion boundary pixels (suppose there are $G_{p}$ pixels above the threshold). The calculated ground truth occlusion boundary maps for each LF data are shown on the right side of Fig. \ref{fig_curvePR}.

To calculate the PR curve of occlusion boundaries, we set different threshold values to the gradients of depth outputs from each algorithm. Suppose the number of pixels above a specific gradient threshold is $C_p$, then the recall rate is $\frac{C_p}{G_p}$. If $T_p$ pixels out of $C_p$ are consistent with the ground truth occlusion boundary pixels, then the precision rate is calculated as $\frac{T_p}{C_p}$.

The average PR curves for all LF data in the HCI dataset by different algorithms are plotted in Fig. \ref{fig_curvePR}. As can be seen, the proposed algorithm has the best occlusion boundary precision at all recall rates, which validates the effectiveness of the proposed method in handling POBR uncertainties and capturing more precise boundary borders.

\begin{table*}[ht]
	\begin{center}		
		\caption{Computation time comparison for different methods on the HCI dataset (in \textit{sec}). For Jeon et al. \cite{jeon2015accurate}, the \textit{Initial} step involves cost volume construction and aggregation. For Wang et al. \cite{wang2016occlusion}, the \textit{Initial} step involves initial label estimation and occlusion analysis. For our proposed method, the \textit{POBR Optimize} step involves confidence map/edge weights refinement and final weighted least squares optimization.} 
		\label{tbl_complexityEval}
		\begin{tabular}{|m{1cm}|>{\centering\arraybackslash}m{1cm}|>{\centering\arraybackslash}m{0.8cm}|>{\centering\arraybackslash}m{1cm}|>{\centering\arraybackslash}m{0.8cm}|>{\centering\arraybackslash}m{0.8cm}|>{\centering\arraybackslash}m{0.8cm}|>{\centering\arraybackslash}m{0.8cm}|>{\centering\arraybackslash}m{1.2cm}|>{\centering\arraybackslash}m{1cm}|>{\centering\arraybackslash}m{1.1cm}|>{\centering\arraybackslash}m{0.8cm}|}
			\hline 
			&\multicolumn{4}{c|}{Jeon et al.} &\multicolumn{3}{c|}{Wang et al.} &\multicolumn{4}{c|}{Proposed} \\ 
			\cline{2-12}
			&\textit{Initial} &\textit{Graph Cut} &\textit{Iterative refine}  & \textbf{Total} &\textit{Initial} &\textit{Graph Cut} &\textbf{Total} &\textit{Pixel-wise Depth} &\textit{SP-wise Depth} &\textit{POBR Optimize} &\textbf{Total} \\
			\cline{2-12}
			&MATLAB &C &MATLAB &-- &C &C &-- &C &MATLAB &MATLAB &-- \\ \hline
			\textit{Buddha} &4088	&634	&2030 &\textbf{6752} &686	&116	&\textbf{802} &406	&140	&4	&\textbf{550} \\ 
			\textit{Buddha2}&4033	&612	&1914 &\textbf{6559} &549	&101	&\textbf{650} &458	&154	&5	&\textbf{616} \\ 
			\textit{Mona} 	&4029	&568	&1706 &\textbf{6303} &767	&144	&\textbf{912} &523	&148	&4	&\textbf{675} \\ 
			\textit{Papillon} &3943	&635	&1822 &\textbf{6400} &702	&125	&\textbf{827} &454	&137	&3	&\textbf{594} \\ 
			\textit{StillLife} &4098	&770	&2283 &\textbf{7151} &891	&167	&\textbf{1058} &513	&154	&4	&\textbf{671} \\ 
			\textit{Horses} &4105 	&633	&1812 &\textbf{6550} &719	&195	&\textbf{914} &456	&148	&4	&\textbf{608} \\ 
			\textit{Medieval} &5005 &752	&2224 &\textbf{7981} &949	&253	&\textbf{1202} &599	&224 &4	&\textbf{827} \\ \hline
			\textbf{Average} &4186	&658	&1970 &\textbf{6814} &752 &157 &\textbf{909} &487 &158 &4 &\textbf{649} \\\hline
		\end{tabular}
	\end{center}
\end{table*}

\begin{figure*}[h]
	\centerline{\subfloat{\includegraphics[width=6.4in]{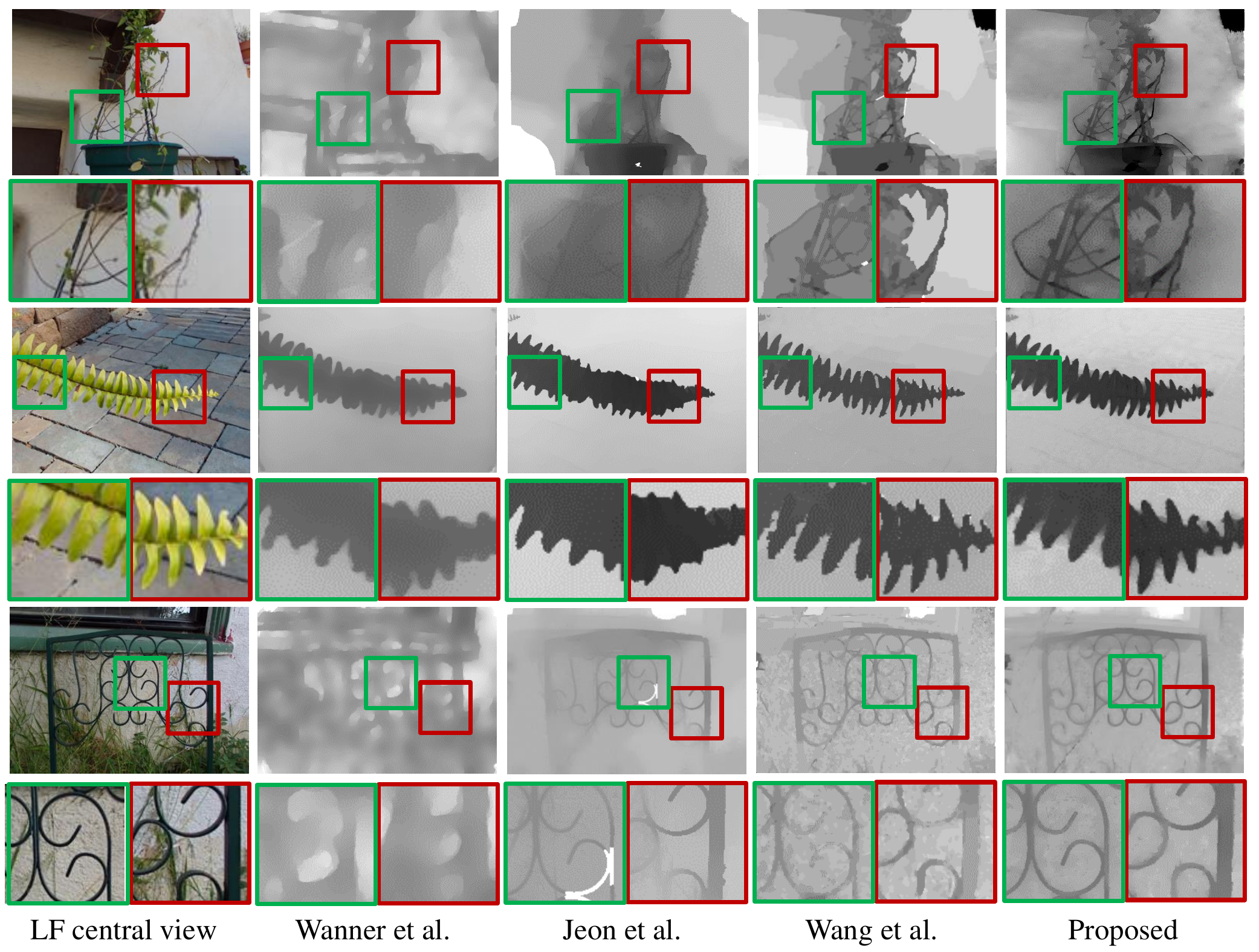}}}
	\caption{Visual comparison of estimated depth maps by Wanner et al. \cite{Wanner2012}, Jeon et al \cite{jeon2015accurate}, Wang et al \cite{wang2016occlusion} and our proposed method for LF data from \cite{wang2016occlusion}. Our method refines occlusion boundaries and keeps intricate structures better than other methods.}
	\label{fig_dataOccu}
\end{figure*}

\begin{figure*}[h]
	\centerline{\subfloat{\includegraphics[width=6.4in]{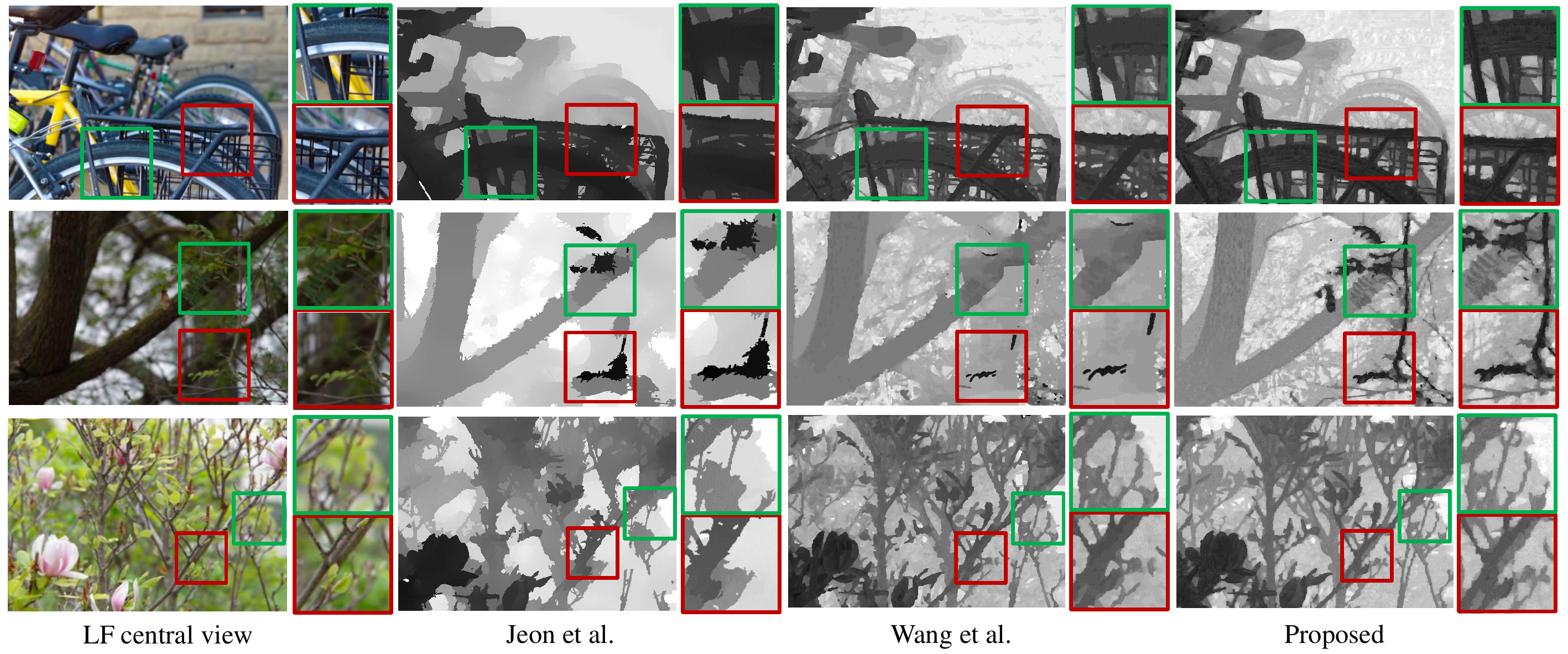}}}
	\caption{Visual comparison of estimated depth maps by Jeon et al. \cite{jeon2015accurate}, Wang et al. \cite{wang2016occlusion} and our proposed method on LF data from the Stanford Lytro LF Archive \cite{stanfordArchive}. Our method shows obvious advantages in areas with multiple occlusions and large depth variations.}
	\label{fig_datastanford}
\end{figure*}

\begin{figure*}[h]
	\centerline{\subfloat{\includegraphics[width=6.4in]{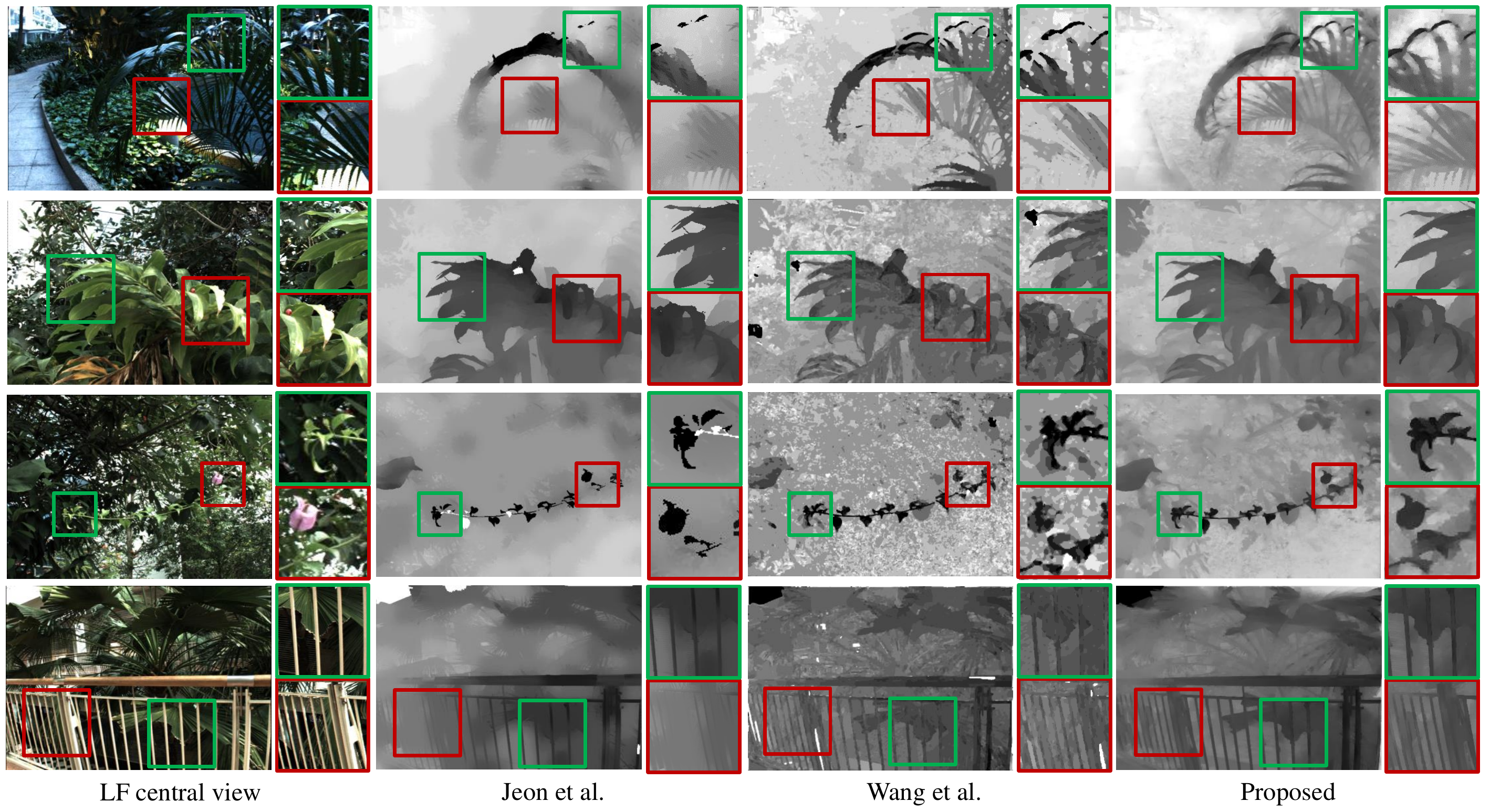}}}
	\caption{Visual comparison of estimated depth maps by Jeon et al \cite{jeon2015accurate}, Wang et al \cite{wang2016occlusion} and our proposed method on LF data captured by our Lytro Illum camera. Our method refines occlusion boundaries and captures structures with fine depth variations better than other methods.}
	\label{fig_dataMine}
\end{figure*}

\subsection{Visual Comparison of Estimated Depth Maps}

We visually evaluate and compare the quality of estimated depth maps by different algorithms. 
First of all, we show the depth estimations for the LF data \textit{Mona} and \textit{Buddha2} from the HCI LF dataset in Fig. \ref{fig_dataHCI}. We can see the depth map from our proposed method is more consistent with the ground truth, especially over occlusion boundaries. Other methods show either underestimated POBR, or noisy predictions around weak textured boundaries. Our estimations over textureless regions are also less noisy as compared to the other methods. 
We went on to evaluate different algorithms on the dataset provided by Wang et al. \cite{wang2016occlusion}, which includes challenging scenes with large depth variations and intricate foreground occlusions. Selected results for different methods are shown in Fig. \ref{fig_dataOccu}. 
The Stanford Lytro LF Archive \cite{stanfordArchive} provides high quality LF data of different object categories with large disparity variations, and it has also been used for comparison. Depth estimation results by different methods are shown in Fig. \ref{fig_datastanford}.
Finally, we tested the algorithms on a set of LF images captured with our own Lytro Illum camera, and the results are shown in Fig. \ref{fig_dataMine}. As can be seen, for all data sources, our algorithm prevails in capturing fine details of occlusion boundaries, intricate structures and overlaid multiple occlusions. Please zoom in on figures for detail comparison.

\subsection{Comparison of Computational Complexity}

We compare the computational complexity of different methods by recording their execution time (in \textit{sec}) for each LF data in the HCI dataset. The results are listed in TABLE \ref{tbl_complexityEval}. All experiments were carried out on a desktop with Intel i7-4790 CPU (2x @3.60GHz), with 16GB RAM. 

In TABLE \ref{tbl_complexityEval}, detailed time costs on major procedures of each method are given. The programming languages are indicated below. Though it is difficult for direct comparison when different programming tools/resources are used, we can still see that our method is faster (avg. 649 \textit{sec}) over the HCI dataset as compared to both Jeon et al. \cite{jeon2015accurate} (avg. 6814 \textit{sec}) and Wang et al. \cite{wang2016occlusion} (avg. 909 \textit{sec}). Especially, our method shows very obvious advantage in the final optimization step, with only 4 \textit{sec} spent on average.

\section{Conclusion and Future Work} \label{sec_conclusion}

In this paper we have proposed a very effective depth estimation framework based on LF images, in which unlike current state-of-the-art methods, focus has been laid on regularizing the initial label confidence map and edge strength weights. Specifically, we first detect POBRs via SP based regularization. Series of shrinkage/reinforcement operations are then applied on the label confidence map and edge strength weights over the POBRs. We have shown that after weight manipulations, even a low-complexity weighted least squares model can produce much better depth estimation than state-of-the-art methods in terms of average disparity error rate, occlusion boundary precision-recall rate, and the preservation of intricate visual features.

Currently, we are only using one superpixel scale to analyze occlusion. This could be problematic when the occluding structures are much smaller than each superpixel. For the future work, we plan to investigate the possibility of multi-scale SP inference \cite{multiscale2016chen}. Information from SPs of different scales could jointly give a better prediction of occlusions and hopefully will handle the scale problem better.

\section*{Acknowledgment}
\addcontentsline{toc}{section}{Acknowledgment}
The research was partially supported by the ST Engineering-NTU Corporate Lab through the NRF corporate lab@university scheme.

\bibliographystyle{IEEEbib}
\bibliography{refs}

\end{document}